\documentclass[]{style}

\definecolor{lastauthor}{RGB}{143, 68, 115}

\usepackage{marvosym}
\usepackage{eso-pic}

\title{Locate Anything in Videos: Rethinking Efficient Generative Spatio-Temporal Video Grounding}

\author[1]{{Hanoona Rasheed}}
\author[1]{{Haania Siddiqui}}
\author[2]{{Ming-Hsuan Yang}}
\author[1,3]{{Fahad Shahbaz Khan}}
\author[1,3]{{Salman Khan}}
\affiliation[1]{Mohamed bin Zayed University of Artificial Intelligence}
\affiliation[2]{University of California, Merced}
\affiliation[3]{Apertix}

\usepackage[table]{xcolor}

\usepackage{amsmath}
\usepackage{amssymb}
\usepackage{graphicx}
\usepackage{booktabs}
\usepackage{xcolor}
\usepackage{hyperref}
\usepackage{cleveref}
\crefname{section}{Sec.}{Secs.}
\Crefname{section}{Sec.}{Secs.}
\usepackage{natbib}
\setcitestyle{numbers,square,sort&compress,citesep={,}}
\usepackage{enumitem}
\usepackage{multirow}
\usepackage{microtype}
\usepackage{tabularx}

\usepackage{makecell}

\hypersetup{
  colorlinks=true,
  linkcolor=metablue,
  citecolor=metablue,
  urlcolor=metablue,
}

\abstract{
Spatio-temporal video grounding (STVG) requires models to identify when a referred event occurs and localize the target entity throughout that interval. Existing multimodal large language models typically serialize dense localization trajectories autoregressively, causing decoding latency to grow with tube length and allowing localization errors to propagate across time. We introduce \textbf{Parallel Tube Decoding (PTD)}, a generative formulation that decomposes grounding into a temporal block followed by time-conditioned spatial blocks decoded simultaneously. This removes both token-level and trajectory-level dependencies, reducing the sequential decoding depth to a fixed $1 + 1$ rounds, independent of tube length. To enable parallel spatial generation, we introduce Decoupled Block Attention, which preserves access to shared video-query context while eliminating cross-box dependencies, together with localization-aware policy optimization for temporal boundaries and spatial geometry. On VidSTG, PTD reduces Tube Completion Latency by $79\times$ and increases spatial decoding throughput by $92\times$ over standard autoregressive decoding, while also improving grounding accuracy. With a compact 4B backbone, our model performs favorably well on VidSTG and HC-STVG, and generalizes zero-shot to temporal grounding, grounded VideoQA, and referring video object tracking. Our results show parallel tube generation is an efficient and effective alternative to autoregressive localization in videos.
}

\newcommand{\projecticon}{\includegraphics[width=1.35em]{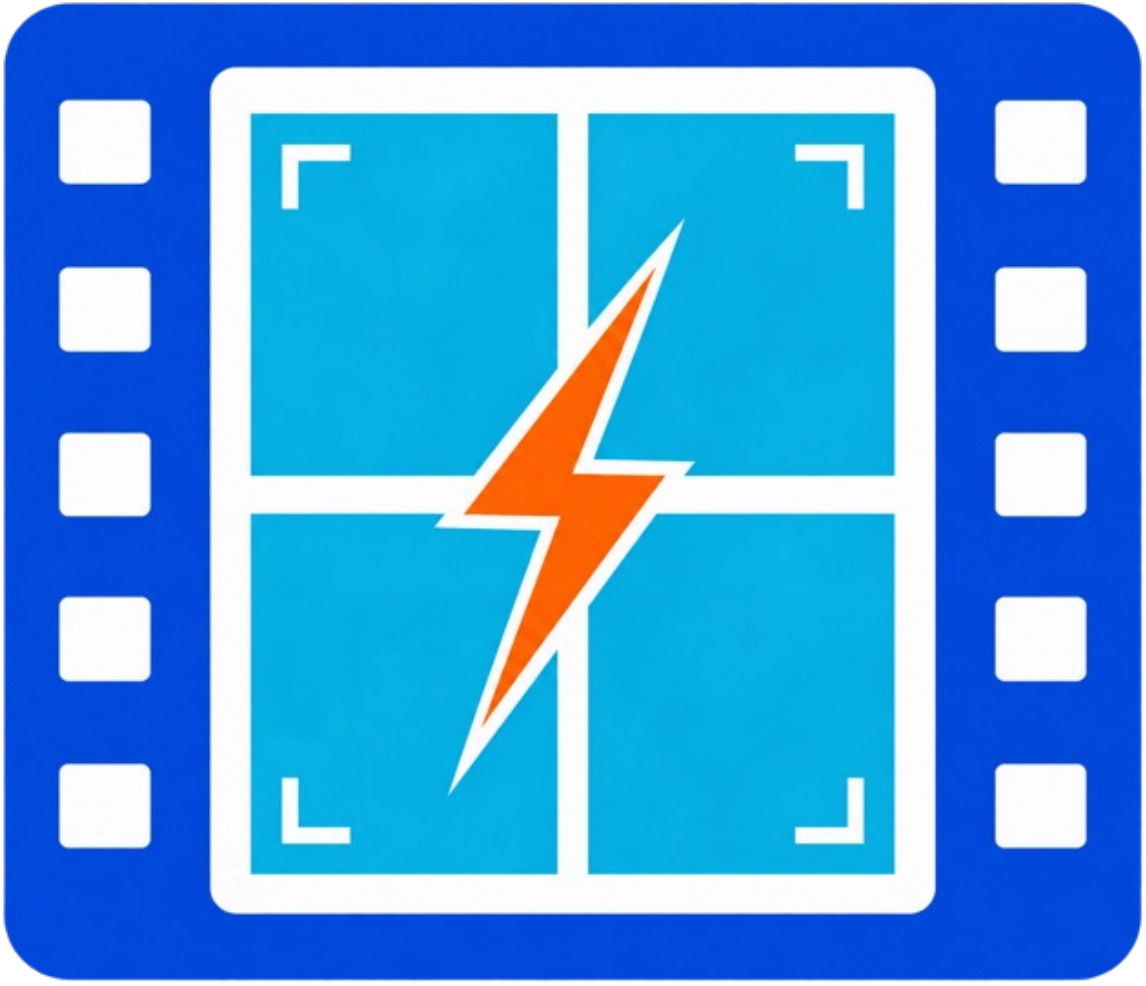}}
\newcommand{\githubresourceicon}{\includegraphics[width=1.35em]{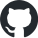}}
\resource{\projecticon}{Website}{https://mbzuai-oryx.github.io/parallel-tube-decoding/}
\resource{\githubresourceicon}{Code}{https://github.com/mbzuai-oryx/ParallelTubeDecoding}

\usepackage[dvipsnames]{xcolor}
\definecolor{GainColor}{named}{RawSienna}
\definecolor{BadColor}{named}{Red}

\usepackage{arydshln}
\usepackage[table]{xcolor}  

\newtcolorbox{appendixpromptbox}[1]{
  enhanced,
  breakable,
  colback=metabg!65!white,
  colframe=metablue!72!black,
  colbacktitle=metablue!14!white,
  coltitle=metafg,
  fonttitle=\small\sffamily\bfseries,
  fontupper=\footnotesize\ttfamily,
  title={#1},
  boxrule=0.5pt,
  arc=3pt,
  left=5pt,
  right=5pt,
  top=4pt,
  bottom=4pt,
  before skip=5pt,
  after skip=6pt,
}
\newtcolorbox{appendixrecordbox}[1]{
  enhanced,
  breakable,
  colback=meituanyellow!16!white,
  colframe=meituanlink!82!black,
  colbacktitle=meituanyellow!34!white,
  coltitle=metafg,
  fonttitle=\small\sffamily\bfseries,
  fontupper=\footnotesize,
  title={#1},
  boxrule=0.5pt,
  arc=3pt,
  left=5pt,
  right=5pt,
  top=4pt,
  bottom=4pt,
  before skip=5pt,
  after skip=6pt,
}

\usepackage{epigraph}
\usepackage[normalem]{ulem}

\usepackage[utf8]{inputenc}         %
\usepackage[T1]{fontenc}            %
\usepackage{url}                    %
\usepackage{booktabs}               %
\usepackage{amsfonts}               %
\usepackage{nicefrac}               %
\usepackage{microtype}              %
\usepackage{algorithm}
\usepackage{algorithmic}
\usepackage{graphicx}
\usepackage[flushleft]{threeparttable}
\usepackage{float}
\usepackage{multirow}
\usepackage{xspace}
\usepackage{enumitem}
\usepackage[font=small]{caption}
\usepackage{autobreak}
\usepackage{sidecap}
\usepackage{wrapfig}
\usepackage[toc, page, header]{appendix}
\usepackage{tikz}
\usepackage{pifont}
\usepackage{mdframed}
\usepackage{colortbl}

\usepackage{listings}
\usepackage{xcolor}
\usepackage{booktabs}
\usepackage{amssymb}
\usepackage{tikz}

\makeatletter
\newcommand{\linkblue}[1]{%
  \begingroup
  \color{\@linkcolor}#1%
  \endgroup
}
\makeatother
\definecolor{darkgray}{gray}{0.35}
\renewcommand{\emph}[1]{\textit{#1}}

\newcommand{\methodterm}[1]{\textcolor{metablue}{#1}}

\newtcolorbox{promptbox}{
  enhanced,
  breakable,
  width=\linewidth,
  colback=metabg,
  colframe=metablue,
  boxrule=0.4pt,
  arc=1.5mm,
  left=2mm,
  right=2mm,
  top=1.5mm,
  bottom=1.5mm,
  before skip=3pt,
  after skip=3pt,
  fontupper=\small\ttfamily\raggedright\sloppy
}

\begin{document}

\maketitle

\vspace{-0.5em}

\begin{figure}[H]
    \centering
    \includegraphics[
        width=\linewidth,
        keepaspectratio
    ]{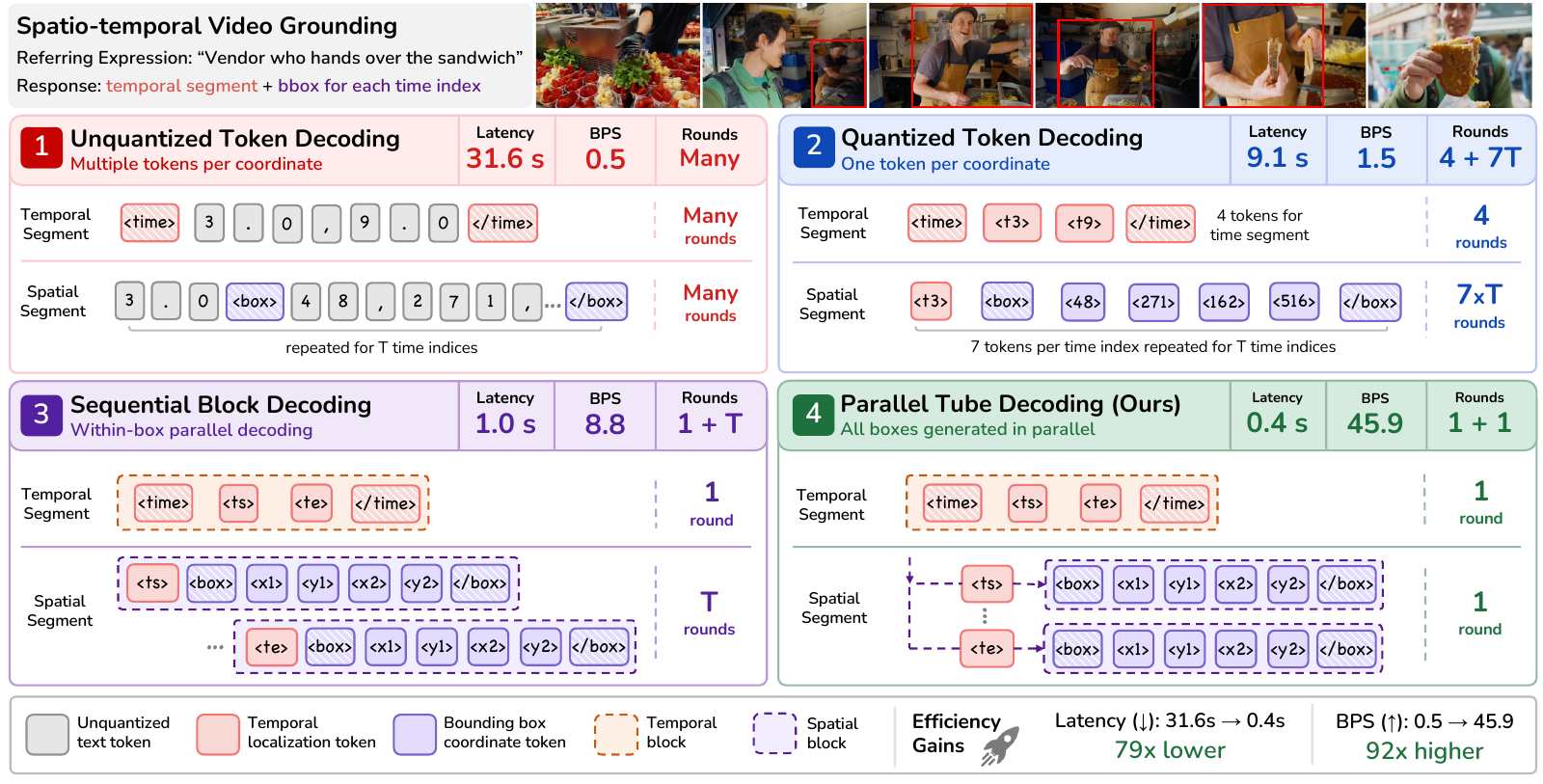}
    \vspace{-20pt}
    \caption{\textbf{Autoregressive localization vs. Parallel Tube Decoding.}
    Given a video and a referring expression, STVG predicts when the event occurs and the bounding box of the referred entity throughout that interval.
    We compare four decoding paradigms. Unquantized Token Decoding generates textual localization values autoregressively, while Quantized Token Decoding reduces representation overhead with discrete \textcolor[RGB]{218,88,82}{temporal} and \textcolor[RGB]{88,24,136}{spatial} tokens. Sequential Block Decoding further removes within-block dependencies but remains sequential across the tube. In contrast, our Parallel Tube Decoding (PTD) generates all time-conditioned spatial blocks in parallel after temporal localization, reducing the sequential decoding depth to $1+1$.
    Compared with standard Unquantized Token Decoding, PTD achieves $79\times$ lower Tube Completion Latency and $92\times$ higher spatial decoding throughput, measured as Boxes Per Second (BPS), while simultaneously improving spatio-temporal grounding accuracy.
    }
    \label{fig:teaser}
\end{figure}

\clearpage

\setlength{\abovedisplayskip}{6pt}
\setlength{\belowdisplayskip}{6pt}
\setlength{\abovedisplayshortskip}{6pt}
\setlength{\belowdisplayshortskip}{6pt}

\section{Introduction}
\label{sec:introduction}
Real-world video understanding requires interpreting natural-language queries about entities and events unfolding across space and time. Spatio-temporal video grounding (STVG) formalizes this by identifying when a queried event occurs and localizing the referred entity throughout that interval~\citep{zhang2020does_vidstg_data,tang2021human_hcstvg_data,yang2022tubedetr}. \textit{Accurate} and \textit{efficient} spatio-temporal localization is becoming increasingly important as video-language models are deployed for long-video search~\citep{wang2024videoagent,ren2024timechat,huang2024vtimellm}, evidence-grounded question answering~\citep{chen2024rextime,chen2025cg,maaz2025video,rasheed2025video}, language-guided tracking~\citep{clark2026molmo2,yuan2025sa2va,bai2024one_reasonvos,munasinghe2025videoglamm}, and embodied interaction~\citep{kim2024openvla,brohan2023rt,chen2025internvla}. Since the query may identify the target through its attributes, actions, or relations rather than a category name, the model must reason over the video to find the intended target, making multimodal large language models (MLLMs) a natural foundation~\citep{bai2025qwen3vl,li2025llavaST,wang2026spacevllm}.

However, STVG remains challenging because temporal and spatial localization involve different sources of ambiguity. Temporally, the target may remain visible before and after the event, while the event itself may be brief or have ambiguous boundaries. Spatially, the target may undergo motion, occlusion, scale and appearance changes, or interference from nearby distractors~\cite{zhang2020does_vidstg_data,tang2021human_hcstvg_data,yang2022tubedetr,gu2024context}. The output structure amplifies this. Unlike temporal grounding, which predicts a start and an end, or image grounding, which predicts a single region, STVG emits a bounding box at every grounded time step. In MLLMs, serializing this tube into an autoregressive coordinate sequence makes decoding depth and latency grow with tube length. 

Because of this cost, many MLLM-based methods offload spatial localization to prediction heads, external detectors, spatial decoders, or detection-and-tracking pipelines~\citep{wang2026spacevllm,devil_gao2025,tu2026bridgestg,stvgr1}; those that keep the tube inside the native token interface stay unified~\citep{stvgo1} but inherit the scaling cost of decoding. The cost can be attacked at the token level, as \Cref{fig:teaser} shows. \methodterm{Unquantized Token Decoding} writes timestamps and coordinates as text, so a single value may be split into multiple tokens and require several dependent predictions. \methodterm{Quantized Token Decoding} maps each value to a single token and substantially reduces this representation overhead~\citep{chen2021pix2seq,jiang2026detect,wang2026locateanything,beyer2024paligemma}. However, it still generates the tokens that describe a bounding box one by one. Inspired by the box-aligned multi-token formulation of~\citep{wang2026locateanything}, we adapt the block decoding to STVG, referring to this formulation as \methodterm{Sequential Block Decoding}. Here, each temporal interval or bounding box is predicted as a single joint block. This removes dependencies within a localization unit but not across them. As a result, the tube is still decoded autoregressively box by box, so decoding depth grows with the tube length.

Removing dependencies inside a box does not remove the dependency between boxes, and this is costly in two ways for STVG. First, it sets a decoding depth that \textit{grows linearly} with tube length, and secondly, it hurts \textit{accuracy}: conditioning each box on an expanding history of generated coordinates lets an early error propagate along the trajectory and shifts the model's attention toward its own output and away from the visual evidence (\Cref{fig:plt_plots}b and c). Neither cost is inherent to the STVG task, since the correct box at a time step is determined by the query and the visual content at that step, not by the boxes emitted before it.

To address these challenges, we introduce \textbf{Parallel Tube Decoding (PTD)}, a generative paradigm for efficient and accurate spatio-temporal video grounding. Our key idea is to decompose tube generation into a temporal block that identifies the queried event and a set of time-conditioned spatial blocks, one per temporal position, that are predicted in parallel across the interval. Tube generation therefore takes two decoding rounds, one temporal and one spatial, regardless of tube length. To enable this while preserving the model's native autoregressive capability, we jointly train complementary next-token and multi-token formulations and introduce {Decoupled Block Attention}, which lets every spatial block access the shared video-query context while blocking dependencies on other predicted boxes. We further add localization-aware policy optimization, improving temporal boundaries and box geometry through complementary spatio-temporal rewards.

The main contributions of this work are:

\begin{itemize}[leftmargin=1.5em, itemsep=0pt, topsep=2pt]

\item \textbf{Decoding formulation.} We study four decoding strategies for generative STVG and introduce Parallel Tube Decoding, reducing tube generation from $4+7T$ rounds under Quantized Token Decoding and $1+T$ under Sequential Block Decoding to just two rounds, independent of the number of boxes (\Cref{sec:method_strategies}).

\item \textbf{Attention mechanism.} We enable PTD through Decoupled Block Attention, which removes cross-box dependencies while preserving shared multimodal evidence, keeping each box grounded in the visual content at its own temporal position (\Cref{sec:method_ptd}).

\item \textbf{Localization-aware optimization.} Policy optimization with complementary temporal and spatial rewards directly improves event boundaries and bounding-box geometry (\Cref{sec:method_grpo}).

\item \textbf{Efficiency and generalization.} PTD cuts Tube Completion Latency by $79\times$ and raises spatial decoding throughput by $92\times$ over unquantized token decoding while improving localization quality (\Cref{sec:exp_strategies}). With a 4B backbone it reaches state-of-the-art or competitive STVG performance (\Cref{sec:exp_stvg}) and generalizes zero-shot to temporal grounding, evidence-grounded VideoQA, and referring video object tracking (\Cref{sec:exp_generalization}).

\end{itemize}
\section{Related Works}
\label{sec:related}

\paragraph{Spatio-Temporal Video Grounding.}
Early STVG approaches rely on task-specific graph reasoning, cross-modal transformers, and DETR-style decoders to jointly infer temporal boundaries and regress the corresponding spatial tube~\citep{zhang2020does_vidstg_data,yang2022tubedetr,gu2024context,gu2025knowing}. Concurrently, MLLMs have demonstrated strong grounding capabilities in static images through coordinate or region generation~\citep{peng2023kosmos,chen2023shikra,ma2024groma}, while video MLLMs extend this capability to temporal localization through timestamp-aware representations and event-boundary prediction~\citep{huang2024vtimellm,ren2024timechat,guo2025vtg}. Recent MLLM-based STVG methods build on these advances along two directions. LLaVA-ST and STVG-o1 directly serialize temporal boundaries and frame-wise boxes through the language decoder, preserving a unified generative interface~\citep{li2025llavaST,stvgo1}. In contrast, many competitive approaches retain the MLLM primarily for semantic reasoning and temporal localization, while delegating dense spatial grounding to dedicated components: SpaceVLLM and Bridge-STG employ specialized spatial decoders~\citep{wang2026spacevllm,tu2026bridgestg}, DEViL couples the MLLM with open-vocabulary detectors~\citep{devil_gao2025}, and STVG-R1 reformulates coordinate prediction as instance-ID selection over preconstructed object trajectories~\citep{stvgr1}. These formulations expose a central trade-off: direct generation retains a unified MLLM output space but produces long serialized tubes, whereas decoupled systems reduce the generation burden by shifting spatial localization to additional modules.

\paragraph{Efficient Localization Decoding.}
This architectural distinction has also shaped how recent STVG methods address inference efficiency. Bridge-STG and DEViL explicitly reduce MLLM output length and decoding cost by offloading dense spatial localization to a dedicated decoder or detector~\citep{tu2026bridgestg,devil_gao2025}. For models that retain localization within the language decoder, discrete spatial and temporal tokens reduce representation overhead, but the complete tube remains restricted by token-by-token generation. Recent advances in multi-token prediction and block-based generation mitigate this limitation by predicting several tokens jointly, although they typically operate on arbitrary token spans or preserve causal dependencies across successive blocks~\citep{liu2025sequential,gloeckle2024better,arriola2025block,nie2026large,zeng2025diffusionvl}. Specifically, LocateAnything treats the entire coordinate set of a bounding box as an atomic prediction block, significantly accelerating visual grounding while preserving geometric coherence~\citep{wang2026locateanything}. Crucially, its box blocks remain causally ordered, such that generating a longer sequence of boxes still requires additional dependent decoding rounds. Based on the above discussion, efficient native STVG generation requires removing not only token-level dependencies within each box, but also the trajectory-level dependency across the complete spatio-temporal tube.

\begin{figure}[t]
    \centering
    \includegraphics[width=0.99\columnwidth]{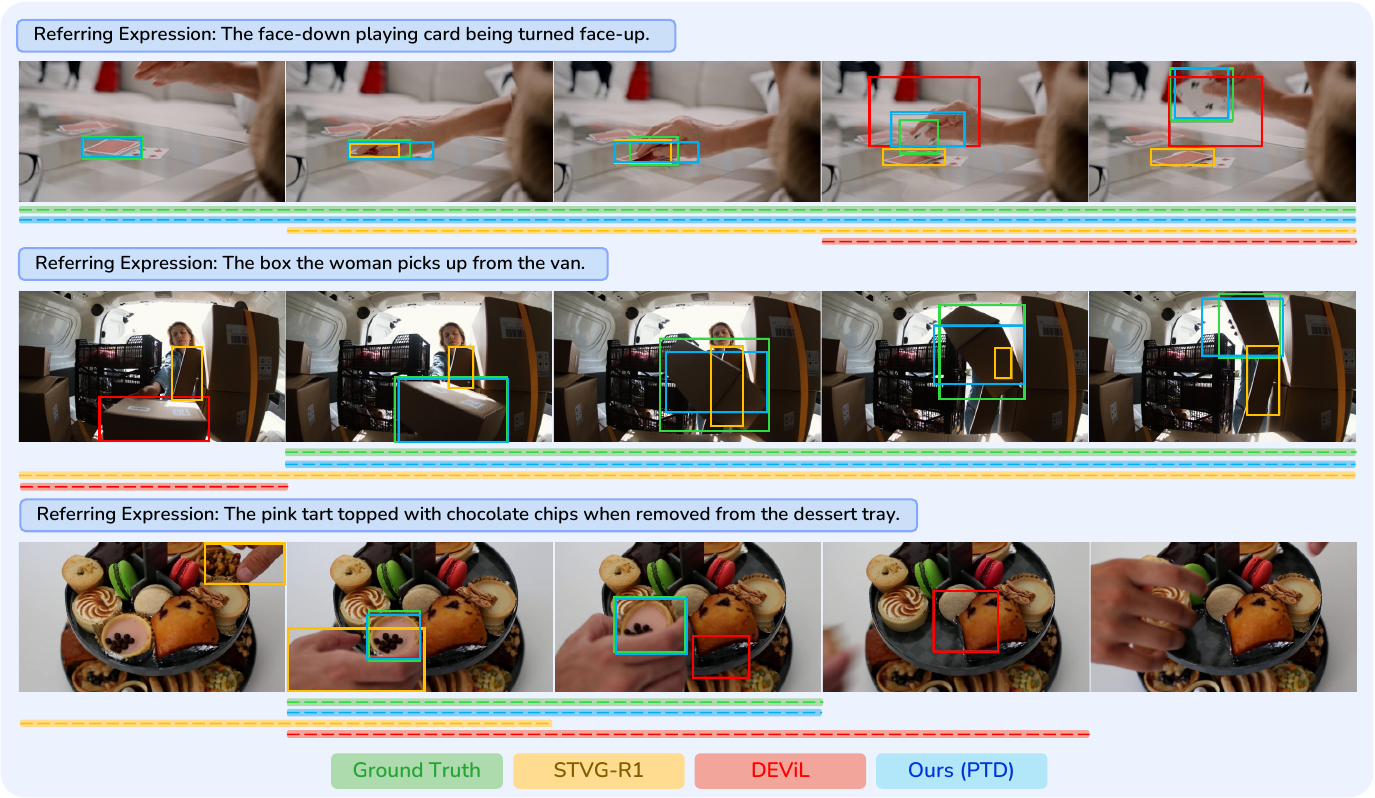}
    \caption{\textbf{Qualitative comparison with prior STVG methods.} We compare our model with competitive STVG methods on challenging grounding examples. Like many existing approaches, STVG-R1~\citep{stvgr1} and DEViL~\citep{devil_gao2025} use the MLLM for temporal localization but construct the spatial tube using external detection or tracking modules. In contrast, PTD directly generates both the temporal interval and complete spatial tube, producing more accurate temporal boundaries and spatial localization. The examples highlight fine-grained target identification among visually similar distractors, large changes in object scale as the target approaches the camera, and event-specific temporal localization rather than simply predicting the full period in which the target is visible. Bounding boxes show spatial predictions, while the horizontal lines indicate the predicted temporal intervals.}
    \label{fig:qualitative_main}
\end{figure}

\section{Method}
\label{sec:method}

In this section, we present our approach to efficient spatio-temporal video grounding (STVG). We begin by formulating the task and then discuss \textit{four} decoding strategies that progressively reduce sequential localization dependencies (\Cref{sec:method_strategies}). To address the remaining sequential dependency, we introduce Parallel Tube Decoding (PTD) (\Cref{sec:method_ptd}). Next, we define latency and throughput metrics for decoding efficiency (\Cref{sec:method_metric}) and present localization-aware policy optimization for improved spatial and temporal grounding (\Cref{sec:method_grpo}).

\textbf{Task Formulation.}\;
Given a video and a referring expression, STVG requires identifying \emph{when} the event described by the query occurs and \emph{where} the referred entity is located throughout that interval. This is challenging as the entity may remain visible beyond the event, while its location can change substantially over time. The model performs temporal localization by predicting the interval $[t_s,t_e]$, and spatial localization by predicting a bounding box $b_i=(x1_i,y1_i,x2_i,y2_i)$ for each time index $t_i$ within that interval. The output is $\mathcal{Y}=([t_s,t_e],\mathcal{B})$, where $\mathcal{B}={(t_i,b_i)}_{i=s}^{e}$ denotes the spatio-temporal tube. We represent the prediction as
\begin{equation}
\texttt{<time>} \;
\texttt{<} t_s \texttt{>} \;
\texttt{<} t_e \texttt{>} \;
\texttt{</time>}
\quad
\left[
\texttt{<} t_i \texttt{>} \;
\texttt{<box>} \;
\texttt{<} x1_i \texttt{>} \;
\texttt{<} y1_i \texttt{>} \;
\texttt{<} x2_i \texttt{>} \;
\texttt{<} y2_i \texttt{>} \;
\texttt{</box>}
\right]_{i=s}^{e}.
\label{eq:response}
\end{equation}

\subsection{From Autoregressive Localization to Parallel Tube Decoding}
\label{sec:method_strategies}

We compare four decoding strategies in terms of \emph{sequential decoding depth}, defined as the number of dependent generation rounds required to produce the complete spatio-temporal tube.

\textbf{i) Unquantized Token Decoding.}\;
Under standard autoregressive decoding, time indices and spatial coordinates are represented as text and generated through the model's language-token prediction~\citep{bai2025qwen3vl,stvgo1}. For example, a timestamp such as \texttt{3.0 seconds} or a coordinate such as \texttt{512} may be decomposed into multiple tokens depending on the tokenizer. Each localization can therefore require \textit{several} sequential decoding steps, resulting in long generation sequences for dense spatio-temporal tubes.

\textbf{ii) Quantized Token Decoding.}\; Quantization addresses this representation overhead by mapping spatial and temporal locations to discrete tokens. Following prior quantized coordinate representations~\citep{chen2021pix2seq,jiang2026detect,wang2026locateanything, beyer2024paligemma, li2025llavaST}, normalized spatial coordinates are quantized into $1{,}001$ bins, such that a coordinate such as $512$ is represented by a single token \texttt{<512>}. For temporal localization, we similarly quantize text timestamps into discrete temporal tokens~\citep{wang2024grounded,li2025llavaST}. Each video temporal patch is prefixed with its corresponding token \texttt{<}$t_i$\texttt{>}, and a time interval is represented as \texttt{<time>} \texttt{<}$t_s$\texttt{>} \texttt{<}$t_e$\texttt{>} \texttt{</time>}, where $t_s$ and $t_e$ denote the start and end positions.

This quantized representation makes the grounding output compact and \textit{fixed in length}, while preserving standard autoregressive decoding. Following the output formulation above, the temporal interval requires \textit{four} sequential token predictions, while each time-indexed box prediction requires \textit{seven}: \texttt{<}$t_i$\texttt{>}, \texttt{<box>}, four coordinate tokens, and \texttt{</box>}. Thus, for a tube containing $T$ boxes, the sequential decoding depth is $4+7T$. Quantization therefore improves token efficiency, but the complete tube is still generated \textit{one token at a time}.

\textbf{iii) Sequential Block Decoding.}\; While quantization reduces token count, decoding remains fully autoregressive. Multi-token prediction (MTP)~\citep{liu2025sequential,zeng2025diffusionvl} reduces this sequential depth by predicting multiple tokens jointly. Following structured block decoding~\citep{wang2026locateanything}, we define each prediction block as a complete localization unit. For STVG, we define two block types: a temporal block, $\texttt{<time>}~\texttt{<}t_s\texttt{>}~\texttt{<}t_e\texttt{>}~\texttt{</time>}$, and a time-indexed spatial block, $\texttt{<}t_i\texttt{>}~\texttt{<box>}~\texttt{<}x1_i\texttt{>}~\texttt{<}y1_i\texttt{>}~\texttt{<}x2_i\texttt{>}~\texttt{<}y2_i\texttt{>}~\texttt{</box>}$. All tokens within a block are predicted jointly in a single decoding round. We use a block size of seven, padding the shorter temporal block with $\texttt{<null>}$ tokens. Each target block is prompted using the final token of the preceding block followed by \texttt{<mask>} tokens, allowing all tokens in the block to be predicted in parallel~\cite{liu2025sequential,wang2026locateanything}.

We refer to this formulation as \emph{Sequential Block Decoding}. Although each block is decoded jointly, they are still generated sequentially along the tube. Thus, a tube with $T$ boxes requires one round for the temporal block and $T$ rounds for spatial localization, reducing the sequential decoding depth from $4+7T$ to $1+T$. This removes token-level dependencies within each block, but leaves a trajectory-level dependency across time. As the tube becomes longer, each additional spatial prediction introduces another sequential decoding round.

\textbf{iv) Parallel Tube Decoding.}\;
We introduce \emph{Parallel Tube Decoding} (PTD) to remove the remaining trajectory-level dependency by predicting all spatial blocks in parallel after temporal localization. The sequential decoding depth therefore becomes $1+1$, independent of tube length $T$. Longer tubes increase only the parallel decoding width, rather than the number of sequential decoding rounds. 

\subsection{Parallel Tube Decoding}
\label{sec:method_ptd}
\begin{wrapfigure}{r}{0.38\columnwidth}
    \centering
    \vspace{-20pt}
    \includegraphics[width=\linewidth]{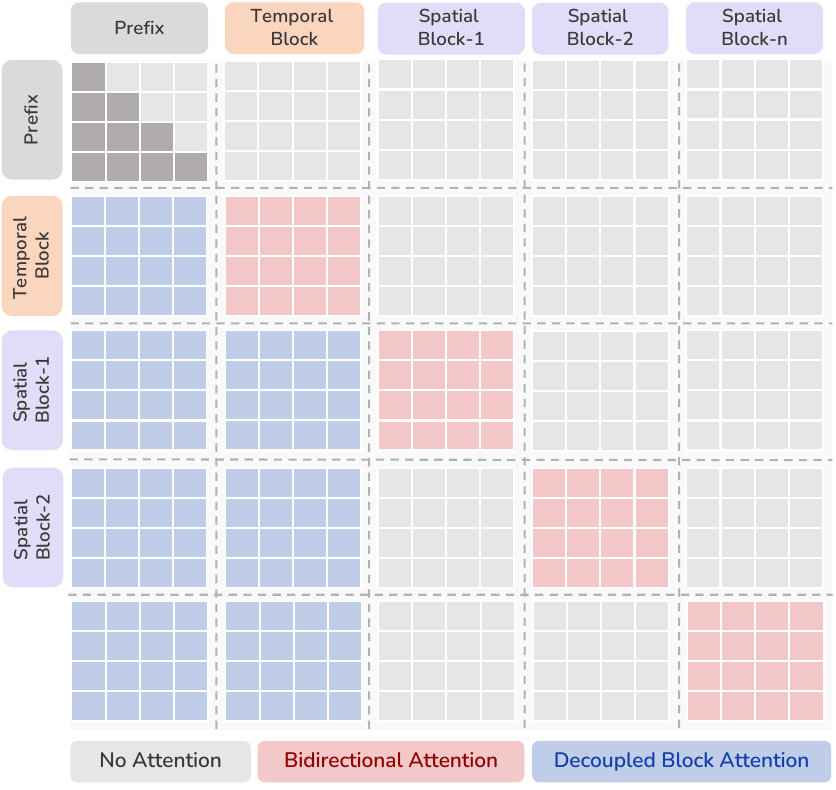}
    \caption{\textbf{Illustration of Decoupled Block Attention in PTD.} Each spatial block attends to the shared multimodal prefix and temporal block, uses bidirectional attention within the block, and remains isolated from other spatial blocks, enabling parallel tube generation.}
    \vspace{-20pt}
    \label{fig:ptd_attention}
\end{wrapfigure}
Sequential block decoding removes token-level dependencies within each localization block, but still generates the spatial trajectory one time step at a time. However, this trajectory-level dependency is not inherent to STVG. Once the temporal interval is localized, spatial grounding at each time index can be performed directly from the corresponding visual evidence, without relying on boxes predicted at earlier time steps. Based on this observation, we introduce \textbf{Parallel Tube Decoding (PTD)}, which decomposes tube generation into two stages: a \textit{temporal block} that identifies the relevant interval, followed by a set of \textbf{time-conditioned spatial blocks} that localize the referred entity throughout that interval. The key here is that all spatial blocks share the same multimodal context, while each is conditioned on its corresponding temporal token. Crucially, all spatial blocks are decoded in parallel, allowing the complete spatial trajectory to be generated in a single round.

Let $P$ denote the shared multimodal prefix, consisting of the video, the referring expression, and the predicted temporal block. Each video temporal patch is associated with a temporal token $\texttt{<}t_i\texttt{>}$, which serves as the \textit{temporal anchor} for its corresponding spatial block. Once the temporal boundaries $[t_s,t_e]$ are predicted, we directly instantiate the corresponding anchors ${\texttt{<}t_i\texttt{>}}_{i=s}^{e}$ from the localized interval rather than generating them autoregressively. Each anchor initializes a spatial block $B_i$ with $\texttt{<}t_i\texttt{>}, \texttt{<mask>}^{5}$ as input, from which the model jointly predicts $\texttt{<box>}, \texttt{<}x1_i\texttt{>}, \texttt{<}y1_i\texttt{>}, \texttt{<}x2_i\texttt{>}, \texttt{<}y2_i\texttt{>}, \texttt{</box>}$. We therefore formulate tube prediction as
\begin{equation}
p(B_{s:e}\mid P)
=
\prod_{i=s}^{e}
p\!\left(B_i \mid P,\texttt{<}t_i\texttt{>}\right).
\label{eq:ptd_factorization}
\end{equation}

To train this parallel decoding formulation while retaining the model's native autoregressive behavior, we adopt a dual-formulation training strategy following~\citep{wang2026locateanything}. Specifically, each training sample is supervised through two complementary target sequences: an NTP sequence that maintains standard causal generation and an MTP sequence that enables structured block prediction, representing the same sequence as a temporal block followed by a set of time-conditioned spatial blocks, $B_s,\ldots,B_e$. Both target sequences are concatenated after the shared video-query context and jointly optimized within a single forward pass: $\mathcal{L}_{\mathrm{SFT}}=\mathcal{L}_{\mathrm{NTP}}+\mathcal{L}_{\mathrm{MTP}}$.

Our attention design combines three attention mechanisms to support parallel tube generation. i) The NTP sequence uses standard causal attention, preserving the model's autoregressive generation ability. ii) Tokens within each MTP block use bidirectional attention, allowing the structured localization unit to be predicted jointly~\citep{wang2026locateanything,fu2025efficient,arriola2025block}. iii) We introduce \emph{\textbf{Decoupled Block Attention}} across time-conditioned spatial blocks. Each spatial block can attend to the shared multimodal prefix, including the predicted temporal block, but is restricted from attending to any other spatial block. Unlike sequential block decoding~\citep{wang2026locateanything}, which uses causal attention across blocks, our design eliminates this cross-block dependency and seamlessly enables the entire spatial trajectory to be decoded in a single generation round.
The NTP causal attention cannot attend to the MTP blocks, preventing information leakage during joint training. 

\subsection{Latency Metrics}
\label{sec:method_metric}
To quantify the decoding efficiency of different decoding strategies, we consider two complementary wall-clock metrics. i) \textbf{Tube Completion Latency (TCL)} measures the time required to complete the spatio-temporal tube prediction once generation begins. Let $T_{\mathrm{full}}$ denote the end-to-end generation time and $T_{\mathrm{TTFT}}$ the time to first token. We define $\mathrm{TCL}=T_{\mathrm{full}}-T_{\mathrm{TTFT}}$, thereby excluding the initial video and prompt processing cost and isolating the latency associated with tube generation. ii) \textbf{Boxes Per Second (BPS)} measures spatial decoding throughput. For a prediction containing $T$ time-conditioned spatial boxes, we define $\mathrm{BPS}=T/\mathrm{TCL}$. These two metrics capture complementary aspects of decoding efficiency: TCL measures how quickly the complete spatio-temporal tube becomes available, while BPS measures the rate at which spatial predictions are produced. Importantly, they expose how each decoding strategy scales with tube length $T$. Token-based decoding and Sequential Block Decoding require increasingly more dependent generation rounds as the tube grows, whereas PTD significantly accelerates tube generation. We evaluate this scaling in~\Cref{sec:exp_strategies}.

\subsection{Localization-Aware Policy Optimization}
\label{sec:method_grpo}
PTD replaces temporally serialized tube generation with parallel time-conditioned spatial prediction. While SFT supervises this formulation through token-level cross-entropy, it does not directly optimize the geometric quality of the resulting tube. We therefore further train the model with localization-aware Group Relative Policy Optimization (GRPO)~\cite{shao2024deepseekmath}, using complementary rewards for temporal and spatial grounding. Recent studies have also explored GRPO for STVG through localization-specific rewards~\citep{stvgo1,stvgr1}.

\textbf{i) Temporal localization reward.}\; A central challenge in STVG is to localize the queried event within a longer temporal window where the referred entity may remain continuously visible. A second challenge arises when the event is temporally brief, making its precise boundaries difficult to localize. Consequently, the model may either fail to localize the event at the correct time or concentrate on a short, visually salient portion while missing earlier or later phases of the interaction. Further, even accurate spatial predictions cannot recover a tube whose temporal boundaries are incorrect. To explicitly optimize this, we use temporal IoU as an interval-level reward between the predicted temporal segment $[\hat{t}_s,\hat{t}_e]$ and the target segment $[t_s,t_e]$,
\begin{equation}
R_{\mathrm{temp}}
=
\operatorname{tIoU}
\left(
[\hat{t}_s,\hat{t}_e],
[t_s,t_e]
\right).
\label{eq:temporal_reward}
\end{equation}

\textbf{ii) Spatial localization reward.}\; Once the relevant event interval is identified, the next challenge is to precisely localize the referred entity within the selected frames. The predicted tube may still contain boxes that are spatially shifted, overly loose, or gradually drift toward nearby objects or background regions. These localization errors require spatial supervision that captures both region-level overlap and coordinate-level precision. We therefore combine generalized IoU~\citep{rezatofighi2019generalized}, which measures the geometric agreement between the predicted and target regions, with an $L_1$ coordinate penalty that further discourages shifted or loosely fitted boxes. Since temporal extent is already optimized by the temporal localization reward, we evaluate spatial quality only over the temporal intersection $\mathcal{I}=[\hat{t}_s,\hat{t}_e]\cap[t_s,t_e]$, where the predicted and target tubes are temporally aligned.
\begin{equation}
R_{\mathrm{spat}}
=
\frac{1}{|\mathcal{I}|}
\sum_{t_i\in\mathcal{I}}
\left[
\operatorname{GIoU}(\hat{b}_i,b_i)
-
\left\|
\hat{b}_i-b_i
\right\|_1
\right].
\label{eq:spatial_reward}
\end{equation}

We combine the temporal and spatial localization rewards into a single objective for policy optimization. Since both components capture complementary aspects of the spatio-temporal tube, we assign them equal weight and define the total reward as $R = R_{\mathrm{temp}} + R_{\mathrm{spat}}$. This jointly encourages accurate temporal localization and precise spatial grounding within the localized interval.
\section{Experiments}
\label{sec:experiments}

\subsection{Experimental Setup}
We build our model on Qwen3-VL-4B~\citep{bai2025qwen3vl} and equip it with discrete spatial and temporal localization tokens for structured spatio-temporal grounding. Specifically, we augment the vocabulary with $1{,}001$ spatial tokens representing normalized coordinates in $[0,1000]$ and $100$ temporal tokens representing discretized timestamps. These temporal tokens are interleaved with the video representation, providing an explicit temporal anchor for each temporal patch and enabling the time-conditioned spatial blocks used by PTD. We use a temporal patch size of $1$ for finer temporal granularity. We initialize the newly introduced localization tokens from semantically aligned Qwen3-VL vocabulary embeddings. 

\textbf{Training stages.}\; We train our model in two stages. We first perform supervised fine-tuning (SFT) using the dual NTP/MTP formulation introduced in \Cref{sec:method_ptd}, enabling the model to support both autoregressive generation and Parallel Tube Decoding (PTD) within the same training stage. We continue training the model with localization-aware Group Relative Policy Optimization (GRPO), using the temporal reward $R_{\text{temp}}$ and spatial reward $R_{\text{spat}}$ defined in \Cref{sec:method_grpo} to further improve localization quality. For both SFT and GRPO, we use LoRA~\citep{hu2021lora} with rank 32 and optimize the LoRA parameters together with the embeddings of the newly introduced localization tokens. For the controlled comparison in \Cref{sec:exp_strategies}, all four variants are trained for $1$ SFT epoch, while the main PTD model is trained for $2$ epochs with all other settings unchanged. SFT is trained with a learning rate of $2\times10^{-5}$ using the dual NTP/MTP formulation, while GRPO uses only the PTD formulation with a learning rate of $5\times10^{-6}$, $\beta=0.04$, and a sampling temperature of 0.9. Videos are sampled at $2$ FPS with at most $64$ frames. 

\textbf{Training data.}\;
For SFT, we combine the training splits of VidSTG~\citep{zhang2020does_vidstg_data} and HC-STVG~\citep{tang2021human_hcstvg_data}. VidSTG provides both declarative and interrogative referring expressions, while HC-STVG-v1 and HC-STVG-v2 further introduce human-centric grounding samples involving attributes, actions, and interactions in multi-person scenes. Together, these datasets yield approximately $90$K SFT samples. For GRPO, rather than uniformly reusing the complete SFT pool, we construct a more informative subset that emphasizes samples with meaningful variation in localization quality. Specifically, for each training sample, the SFT model generates eight candidate responses, which are evaluated using temporal IoU (tIoU) and video IoU (vIoU). We preferentially retain samples where the sampled responses exhibit substantial within-group variation in temporal or spatio-temporal correctness, providing a stronger relative learning signal for policy optimization~\citep{yu2026dapo}. In contrast, samples that are consistently solved offer limited supervision, while those for which all rollouts fail provide little discrimination among candidate responses. Based on this criterion, we construct a $16$K GRPO training set.

\textbf{Evaluation protocol.}\;
We evaluate our approach along \textit{three} complementary dimensions: \textbf{(i) decoding paradigm} (\Cref{sec:exp_strategies}), where we examine how progressively reducing sequential decoding dependencies affects grounding quality and generation efficiency; \textbf{(ii) in-domain STVG performance} (\Cref{sec:exp_stvg}), where we compare our model with prior methods under the standard STVG evaluation setting; and \textbf{(iii) generalization beyond STVG} (\Cref{sec:exp_generalization}), where we assess zero-shot transfer to temporal grounding, grounded VideoQA, and video object tracking. 
\textbf{Decoding Efficiency.}\; To characterize the efficiency of different localization strategies, we use the metrics introduced in \Cref{sec:method_metric}: Tube Completion Latency (TCL) to measure the time required to complete the predicted tube and Boxes Per Second (BPS) to quantify spatial decoding throughput. For all efficiency evaluations, we use batch size $1$ and BF16 inference on a single 64-GB AMD Instinct MI210 GPU using PyTorch 2.7 and ROCm 6.3. We synchronize GPU operations and measure decode-only time from the first output token to tube completion, excluding multimodal prefill, under the same hardware and runtime.

\begin{table*}[t]
\centering
\small
\setlength{\tabcolsep}{5pt}

\resizebox{\textwidth}{!}{%
\begin{tabular}{lcccccccccc}
\toprule
& \multicolumn{4}{c}{\textbf{Declarative}}
& \multicolumn{4}{c}{\textbf{Interrogative}}
& \multicolumn{2}{c}{\textbf{Efficiency}} \\

\cmidrule(lr){2-5}
\cmidrule(lr){6-9}
\cmidrule(lr){10-11}

Model Design
& $m_{\mathrm{tIoU}}$
& \multicolumn{3}{c}{$\mathrm{vIoU}$}
& $m_{\mathrm{tIoU}}$
& \multicolumn{3}{c}{$\mathrm{vIoU}$}
& TCL (s)
& BPS \\

\cmidrule(lr){3-5}
\cmidrule(lr){7-9}

&
&
Mean
& $@0.3$
& $@0.5$
&
&
Mean
& $@0.3$
& $@0.5$
& $\downarrow$
& $\uparrow$ \\

\midrule

Unquantized Token Decoding
& 42.8 & 27.4 & 39.8 & 27.1
& 40.8 & 22.6 & 32.0 & 20.9
& 31.6 & 0.5 \\

Quantized Token Decoding
& 43.3 & 28.7 & 40.6 & 27.3
& 42.0 & 23.0 & 32.2 & 21.2
& 9.1 & 1.5 \\

Sequential Block Decoding
& 47.2 & 31.2 & 42.8 & 28.7
& 46.3 & 26.0 & 35.5 & 22.5
& 1.0 & 8.8 \\

\midrule

\textbf{Parallel Tube Decoding (PTD)}
& \textbf{47.5}
& \textbf{32.9}
& \textbf{46.0}
& \textbf{30.5}
& \textbf{46.7}
& \textbf{27.2}
& \textbf{37.8}
& \textbf{23.3}
& \textbf{0.4}
& \textbf{45.9} \\

\bottomrule
\end{tabular}%
}

\caption{
We report spatio-temporal grounding performance for declarative and interrogative queries, together with decoding efficiency measured by tube completion latency (TCL) and boxes per second (BPS). Quantization reduces decoding cost, sequential block prediction further improves efficiency, and the proposed \textbf{Parallel Tube Decoding (PTD)} achieves the strongest grounding performance while substantially reducing TCL and increasing BPS.
}
\label{tab:vidstg_efficiency}
\end{table*}
\subsection{Evaluating the Decoding Paradigm}
\label{sec:exp_strategies}
\begin{figure}[t]
    \centering
    \includegraphics[width=0.99\columnwidth]{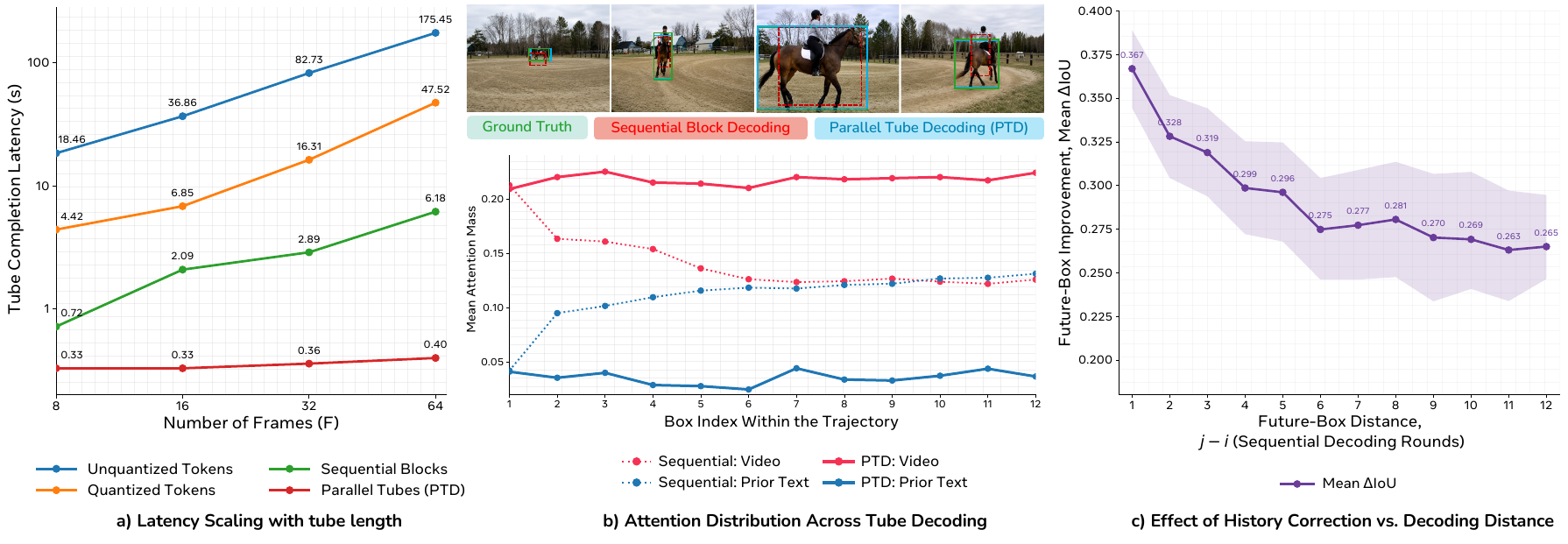}
    \caption{
    \textbf{Analysis of decoding efficiency and trajectory-level dependency.}
    (a) Tube completion latency as the number of grounded frames increases. PTD maintains nearly constant latency, while token-based and block decoding scale with tube length.
    (b) Attention distribution across tube decoding for Sequential Block Decoding (dotted) and PTD (solid). Sequential decoding progressively shifts attention from the video toward prior text.
    (c) History-correction analysis for Sequential Block Decoding. Replacing an erroneous box $B_i$ with its ground-truth box improves subsequent predictions, with the effect gradually decreasing as the decoding distance $j-i$ increases.
    }
    \label{fig:plt_plots}
\end{figure}

\begin{table*}[t]
\centering
\small
\setlength{\tabcolsep}{3.5pt}

\resizebox{\textwidth}{!}{%
\begin{tabular}{lccccccccc}
\toprule
& &
\multicolumn{4}{c}{\textbf{Declarative Sentences}}
& \multicolumn{4}{c}{\textbf{Interrogative Sentences}} \\
\cmidrule(lr){3-6}
\cmidrule(lr){7-10}

\textbf{Model}
& \textbf{Scale}
& $m_{\mathrm{tIoU}}$
& $m_{\mathrm{vIoU}}$
& $\mathrm{vIoU}@0.3$
& $\mathrm{vIoU}@0.5$
& $m_{\mathrm{tIoU}}$
& $m_{\mathrm{vIoU}}$
& $\mathrm{vIoU}@0.3$
& $\mathrm{vIoU}@0.5$ \\

\midrule
\multicolumn{10}{l}{{\scriptsize\linkblue{\textit{Backbone Baselines}}}} \\

Qwen2.5-VL-7B (ZS)~\citep{bai2025qwen25vl}
& 7B
& 16.8 & 10.9 & 14.3 & 5.4
& 13.8 & 8.5 & 11.3 & 4.4 \\

Qwen2.5-VL-7B + SFT$^\dagger$~\citep{stvgo1}
& 7B
& 41.6 & 20.3 & 26.1 & 15.4
& 40.9 & 17.1 & 17.6 & 13.9 \\

Qwen3-VL-4B (ZS)~\citep{bai2025qwen3vl}
& 4B
& 36.2 & 13.1 & 16.6 & 7.0
& 36.1 & 8.9 & 10.2 & 3.8 \\

Qwen3-VL-8B (ZS)~\citep{bai2025qwen3vl}
& 8B
& 37.0 & 13.4 & 16.5 & 7.1
& 35.0 & 9.3 & 11.0 & 3.9 \\

Qwen3-VL-8B + SFT$^\ddagger$~\citep{devil_gao2025}
& 8B
& 42.8 & 22.4 & 29.6 & 15.3
& 41.4 & 18.2 & 19.6 & 12.8 \\

\midrule
\multicolumn{10}{l}{{\scriptsize\linkblue{\textit{Prior MLLM Methods}}}} \\

GroundingGPT~\citep{li2024groundinggpt}
& 7B
& 15.5 & 12.3 & 13.2 & 4.1
& 11.9 & 8.7 & 9.6 & 2.9 \\

Gemini-2.5-Pro~\citep{comanici2025gemini}
& --
& 49.9 & 22.5 & 33.6 & 12.0
& 45.4 & 13.7 & 17.3 & 8.1 \\

LLaVA-ST~\citep{li2025llavaST}
& 7B
& 45.5 & 24.8 & 36.0 & 22.9
& 43.2 & 20.0 & 28.1 & 17.5 \\

SpaceVLLM-7B~\citep{wang2026spacevllm}
& 7B
& 47.7 & 27.4 & 39.1 & 26.2
& 48.5 & 25.4 & 35.9 & 22.2 \\

DEViL (VideoLLaMA3-7B)~\citep{devil_gao2025}
& 7B
& 50.2 & 33.6 & 46.5 & 34.0
& 48.5 & 28.8 & 38.7 & 28.2 \\

STVG-o1 (Qwen2.5-VL-7B)~\citep{stvgo1}
& 7B
& 52.1 & 33.5 & 48.4 & 32.0
& \underline{50.5} & 27.9 & 39.8 & 26.0 \\

Bridge-STG (Qwen3-VL-7B)~\citep{tu2026bridgestg}
& 7B
& \underline{52.6} & \underline{37.2}
& \underline{52.4} & \underline{37.4}
& 50.1 & \underline{31.3}
& \underline{43.8} & \textbf{31.2} \\

\midrule
\multicolumn{10}{l}{{\scriptsize\linkblue{\textit{Ours (Qwen3-VL-4B)}}}} \\

SFT (NTP)
& 4B
& 46.3 & 31.6 & 45.1 & 30.5 & 
44.7 & 26.3 & 36.9 & 25.3 \\

SFT (PTD)
& 4B
& 50.0 & 34.9 & 48.5 & 33.3 & 
48.2 & 28.7 & 40.2 & 25.6 \\

\textbf{GRPO (PTD)}
& {4B}
& \textbf{53.7}
& \textbf{38.3}
& \textbf{53.0}
& \textbf{37.5}
& \textbf{52.2}
& \textbf{32.4}
& \textbf{44.9}
& \underline{30.3} \\

\bottomrule
\end{tabular}%
}

\caption{
Comparison with prior work on \textbf{VidSTG} under declarative and interrogative settings.
We report the model scale, mean temporal IoU ($m_{\mathrm{tIoU}}$), mean video IoU
($m_{\mathrm{vIoU}}$), and $\mathrm{vIoU}$ at 0.3 and 0.5 thresholds.
$^\dagger$ denotes the Qwen2.5-VL SFT baseline reported in STVG-o1~\citep{stvgo1}, while
$^\ddagger$ denotes the Qwen3-VL SFT baseline reported in DEViL~\citep{devil_gao2025}.
Our results use Qwen3-VL-4B as the backbone.
\textbf{SFT (NTP)} uses supervised fine-tuning with standard next-token
autoregressive decoding, whereas \textbf{SFT (PTD)} uses the same supervised
training objective with Parallel Tube Decoding at inference.
\textbf{GRPO (PTD)} further optimizes the model with GRPO while
retaining Parallel Tube Decoding.
}
\label{tab:vidstg_results}
\end{table*}

We first evaluate how the decoding paradigm affects both grounding quality and generation efficiency. In \Cref{tab:vidstg_efficiency}, we compare the four decoding strategies introduced in \Cref{sec:method_strategies} on VidSTG~\citep{zhang2020does_vidstg_data} and report $\mathrm{tIoU}$ and $\mathrm{vIoU}$ for grounding quality. This comparison reveals a clear progression as sequential dependencies are reduced: Unquantized Token Decoding generates textual localization values autoregressively, Quantized Token Decoding replaces each value with a discrete token, Sequential Block Decoding jointly predicts structured localization blocks while remaining sequential across time, and our Parallel Tube Decoding (PTD) removes this remaining dependency by decoding all time-conditioned spatial blocks in parallel after temporal localization. Here, the unquantized variant uses the Qwen3-VL baseline decoding scheme. We adapt existing decoding strategies for STVG to build the quantized and block decoding variants, and propose PTD on top of them. For fair comparison, all variants are trained under the same setting and differ in output representation and decoding strategy used at inference. Across the four variants, the mean predicted tube lengths are $12.3$, $13.7$, $14.2$, and $17.7$ boxes, respectively.
Sequential Block Decoding serves as the natural ablation of Decoupled Block Attention. Replacing the decoupled attention mask with standard causal inter-block attention allows each spatial block to attend to preceding box blocks, thereby restoring sequential dependencies across time and preventing parallel block decoding.

The results show that progressively reducing sequential decoding dependencies improves both grounding quality and generation efficiency. Quantization provides an immediate efficiency gain over unquantized, reducing TCL from $31.6$s to $9.1$s while preserving comparable grounding quality. Block decoding further improves both efficiency and localization performance, reaching $1.0$s TCL and $8.8$ BPS while consistently improving $\mathrm{tIoU}$ and $\mathrm{vIoU}$ across both query types. A key advantage of PTD is that it removes the remaining trajectory-level dependency without sacrificing grounding quality. PTD achieves the strongest performance across all grounding metrics, while further reducing TCL to $0.4$s and increasing throughput to $45.9$ BPS. Compared with the Qwen3-VL-4B baseline using Unquantized Token Decoding~\citep{bai2025qwen3vl}, PTD reduces tube completion latency by $79\times$ and increases spatial decoding throughput by $92\times$, demonstrating that parallel tube generation substantially accelerates inference while simultaneously improving spatio-temporal grounding.

We further study \textbf{latency scaling with tube length}. In \Cref{fig:plt_plots}, we vary the number of boxes from $8$ to $64$ to examine how each decoding strategy scales as the spatial trajectory becomes longer. The latency of token-based and Sequential Block Decoding increases substantially with tube length, reflecting the growing number of dependent generation rounds. In contrast, PTD exhibits nearly constant generation time, increasing only from $0.33$s to $0.40$s as the tube length grows by $8\times$, whereas block decoding rises substantially from $0.72$s to $6.18$s over the same range. Notably, this advantage becomes increasingly pronounced for longer tubes, where PTD avoids the latency growth incurred by sequential decoding. These results demonstrate the key scaling advantage of PTD: longer trajectories increase parallel decoding width rather than sequential decoding depth, enabling efficient tube generation with nearly tube-length-independent latency.

\textbf{Cross-box dependency and error propagation.}\; We investigate why removing trajectory-level dependencies leads to stronger spatial localization. A key observation is that the two decoding strategies exhibit different attention behavior as tube generation progresses. \Cref{fig:plt_plots}b illustrates a qualitative example where block decoding progressively shifts attention from the video toward the growing localization history (dotted lines). In contrast, PTD maintains the video attention (solid lines). We directly quantify the resulting error propagation in  \Cref{fig:plt_plots}c, through a history-correction intervention. For a mislocalized source box $B_i$, we replace only $B_i$ with its ground-truth box and autoregressively re-decode each subsequent box $B_j$. The x-axis $j-i$ measures the number of sequential decoding rounds between the corrected and evaluated boxes, while the y-axis reports the resulting mean $\Delta\mathrm{IoU}$ relative to decoding with the original history. Correcting $B_i$ produces the largest improvement for nearby future boxes, with the effect gradually decreasing with decoding distance. 

\begin{table*}[t]
\centering
\small
\setlength{\tabcolsep}{3.5pt}

\resizebox{\textwidth}{!}{%
\begin{tabular}{lccccccccc}
\toprule
& &
\multicolumn{4}{c}{\textbf{HC-STVG v1}}
& \multicolumn{4}{c}{\textbf{HC-STVG v2}} \\
\cmidrule(lr){3-6}
\cmidrule(lr){7-10}

\textbf{Model}
& \textbf{Scale}
& $m_{\mathrm{tIoU}}$
& $m_{\mathrm{vIoU}}$
& $\mathrm{vIoU}@0.3$
& $\mathrm{vIoU}@0.5$
& $m_{\mathrm{tIoU}}$
& $m_{\mathrm{vIoU}}$
& $\mathrm{vIoU}@0.3$
& $\mathrm{vIoU}@0.5$ \\
\midrule

\multicolumn{10}{l}{{\scriptsize\linkblue{\textit{Backbone Baselines}}}} \\

Qwen2.5-VL-7B (ZS)~\citep{bai2025qwen25vl}
& 7B
& 25.6 & 19.1 & 20.2 & 12.6
& 22.9 & 13.0 & 15.6 & 6.4 \\

Qwen2.5-VL-7B (SFT$^\dagger$)~\citep{stvgo1}
& 7B
& 53.5 & 28.6 & 45.2 & 21.9
& 55.3 & 26.5 & 38.6 & 20.2 \\

Qwen3-VL-4B (ZS)~\citep{bai2025qwen3vl}
& 4B
& 44.6 & 19.5 & 25.4 & 4.9
& 45.2 & 19.4 & 24.7 & 5.5 \\

Qwen3-VL-8B (ZS)~\citep{bai2025qwen3vl}
& 8B
& 47.6 & 21.5 & 30.3 & 6.5
& 53.1 & 21.9 & 30.0 & 6.6 \\

\midrule
\multicolumn{10}{l}{{\scriptsize\linkblue{\textit{Prior MLLM Methods}}}} \\

GPT-4o~\citep{openai2024gpt4o}
& --
& 27.5 & 7.9 & 4.0 & 0.3
& 32.7 & 9.1 & 5.7 & 0.0 \\

Gemini-2.5-Pro~\citep{comanici2025gemini}
& --
& 55.1 & 25.9 & 39.1 & 9.9
& 60.4 & 24.5 & 34.6 & 9.6 \\

GroundingGPT~\citep{li2024groundinggpt}
& 7B
& 22.2 & 16.7 & 15.0 & 4.9
& 19.6 & 14.7 & 16.6 & 3.1 \\

LLaVA-Video SFT~\citep{zhang2024llava}
& 7B
& 52.8 & 27.7 & 43.1 & 21.3
& 54.2 & 24.8 & 40.1 & 15.5 \\

SpaceVLLM~\citep{wang2026spacevllm}
& 7B
& 56.9 & 39.3 & 66.6 & 36.9
& 58.0 & 34.0 & 56.9 & 24.7 \\

STVG-R1 (Qwen2.5-VL-7B)~\citep{stvgr1}
& 7B
& 56.9 & 39.1 & 66.7 & 38.6
& 62.0 & 40.2 & 67.8 & 38.8 \\

DEViL (VideoLLaMA3-7B)~\citep{devil_gao2025}
& 7B
& 59.0 & 43.1 & 70.5 & \underline{44.3}
& 61.7 & \underline{42.5} & 67.3 & \underline{42.2} \\

Bridge-STG (Qwen3-VL-7B)~\citep{tu2026bridgestg}
& 7B
& -- & -- & -- & --
& \textbf{64.1} & 41.5 & 67.5 & 38.6 \\

STVG-o1 (Qwen2.5-VL-7B)~\citep{stvgo1}
& 7B
& \textbf{60.3} & \underline{44.1} & \underline{73.3} & 43.5
& 63.8 & 41.2 & \underline{68.5} & 39.6 \\

\midrule
\multicolumn{10}{l}{{\scriptsize\linkblue{\textit{Ours (Qwen3-VL-4B)}}}} \\
SFT (NTP)
& 4B
& 55.2 & 41.0 & 66.5 & 41.7
& 54.4 & 36.5 & 60.3 & 29.7 \\

SFT (PTD)
& 4B
& 55.7 & 42.7 & 68.6 & 42.5
& 57.0 & 39.2 & 64.7 & 34.0 \\

\textbf{GRPO (PTD)}
& {4B}
& \underline{59.4}
& \textbf{45.6}
& \textbf{73.8}
& \textbf{46.7}
& \underline{63.1}
& \textbf{43.6}
& \textbf{71.0}
& \textbf{42.2} \\

\bottomrule
\end{tabular}%
}

\caption{
Comparison with prior work on \textbf{HC-STVG v1} and \textbf{HC-STVG v2}.
We report the model scale, mean temporal IoU ($m_{\mathrm{tIoU}}$), mean video IoU
($m_{\mathrm{vIoU}}$), and $\mathrm{vIoU}$ at thresholds 0.3 and 0.5.
$^\dagger$ denotes the Qwen2.5-VL SFT baseline reported in STVG-o1.
Our models use Qwen3-VL-4B as the backbone.
\textbf{SFT (NTP)} uses standard next-token prediction, while
\textbf{SFT (PTD)} uses Parallel Tube Decoding.
\textbf{GRPO (PTD)} further optimizes the PTD model with GRPO.
}
\label{tab:hcstvg_results}
\end{table*}
\subsection{Main Results on Spatio-Temporal Video Grounding}
\label{sec:exp_stvg}

We evaluate the spatio-temporal grounding performance of our model on VidSTG~\citep{zhang2020does_vidstg_data} and HC-STVG-v1/v2~\citep{tang2021human_hcstvg_data}, comparing it with backbone baselines and prior STVG methods.

\textbf{i) VidSTG.}\; As shown in \Cref{tab:vidstg_results}, our model establishes the strongest overall performance on VidSTG, leading on seven of eight metrics across declarative and interrogative queries. It improves the previous best $m_{\mathrm{tIoU}}$ by $+1.1$ and $+1.7$ points for declarative and interrogative queries, respectively. The gains are even more pronounced when evaluating the full spatio-temporal tube, which requires both accurate temporal boundaries and consistent spatial localization throughout the event. Our model improves the previous best $m_{\mathrm{vIoU}}$ by $+1.1$ points for both query types, while remaining consistently strong under stricter tube-overlap thresholds.

\textbf{ii) HC-STVG.}\; A similar advantage extends to the more challenging human-centric HC-STVG benchmarks, as shown in \Cref{tab:hcstvg_results}. While temporal localization remains competitive with the best prior methods, the gains are more pronounced when evaluating the full spatio-temporal tube. Our model improves the previous best $m_{\mathrm{vIoU}}$ by $+1.5$ points on HC-STVG-v1 and $+1.1$ points on HC-STVG-v2. This advantage is also reflected under stricter overlap criteria, where it achieves the best $\mathrm{vIoU}@0.3$ on both benchmarks.

The progression within our model further clarifies where these gains originate. SFT (NTP) and SFT (PTD) correspond to two inference modes of the same checkpoint, jointly trained with the NTP and MTP formulations. The former uses Quantized Token Decoding, while the latter uses Parallel Tube Decoding. As already shown in~\Cref{tab:vidstg_efficiency}, PTD improves both grounding quality and decoding efficiency over NTP. Building on this, localization-aware policy optimization further improves grounding quality while retaining the same PTD formulation. A key advantage is that these gains are achieved with a compact Qwen3-VL-4B backbone that directly generates both the temporal interval and the complete spatial tube. 
In contrast, several competitive methods rely on additional spatial components: SpaceVLLM uses a specialized prediction head~\citep{wang2026spacevllm}, DEViL an external detector~\citep{devil_gao2025}, Bridge-STG a dedicated spatial decoder~\citep{tu2026bridgestg}, and STVG-R1 a detection-and-tracking pipeline~\citep{stvgr1}.
Despite using a smaller backbone and no dedicated spatial localization module, our model remains competitive with or surpasses 7B-scale models. Further, the substantial gap over zero-shot Qwen3-VL-4B/8B and existing Qwen3-VL SFT baselines shows that these improvements cannot be attributed to the backbone alone, but are instead driven by the combination of the proposed grounding formulation and localization-aware policy optimization.
\subsection{Generalization Beyond STVG}
\label{sec:exp_generalization}
In this section, we examine whether the learned grounding capabilities transfer beyond the training distribution across three settings: temporal grounding, evidence-grounded VideoQA, and referring video object tracking.

\begin{table*}[t]
\centering
\begin{minipage}[t]{0.48\textwidth}
\centering

{\small
\setlength{\tabcolsep}{6pt}
\resizebox{\textwidth}{!}{%
\begin{tabular}{lcccc}
\toprule
& \multicolumn{4}{c}{\textbf{Charades-STA}} \\
\cmidrule(lr){2-5}
\textbf{Model}
& $\mathrm{R@0.3}$
& $\mathrm{R@0.5}$
& $\mathrm{R@0.7}$
& \textbf{mIoU} \\
\midrule
TRACE-7B~\citep{guo2024trace}       & --   & 40.3 & 19.4 & --   \\
TimeSuite-7B~\citep{zeng2025timesuite}   & 69.9 & 48.7 & 24.0 & --   \\
DEViL~\citep{devil_gao2025}          & 72.6 & 51.5 & 25.2 & 47.7 \\
STVG-R1-7B~\citep{stvgr1}     & 73.2 & 52.5 & --   & --   \\
\midrule
\textbf{Ours}
& \textbf{78.9}
& \textbf{58.4}
& \textbf{30.1}
& \textbf{52.4} \\

\bottomrule
\end{tabular}%
}
}
\captionof{table}{
\textbf{Temporal grounding on Charades-STA.}
We compare all models in the zero-shot setting, evaluating how accurately they localize query-relevant temporal segments at different IoU thresholds.}
\label{tab:charades_sta}
\end{minipage}
\hfill
\begin{minipage}[t]{0.48\textwidth}
\centering

\resizebox{\textwidth}{!}{%
\begin{tabular}{lcccc}
\toprule
& \multicolumn{4}{c}{\textbf{ActivityNet Captions}} \\
\cmidrule(lr){2-5}
\textbf{Model}
& $\mathrm{R@0.3}$
& $\mathrm{R@0.5}$
& $\mathrm{R@0.7}$
& \textbf{mIoU} \\
\midrule
UniTime-Zero~\citep{li2026universal}      & -- & 22.8 & 14.1 & 27.3 \\
Momentor~\citep{qian2024momentor}          & --  & 23.0 & 12.4 & 29.3 \\
Qwen2.5-VL-7B~\citep{bai2025qwen25vl}     & 25.5 & 13.4 & 6.1 & 19.1 \\
VideoChat-TPO~\citep{yan2025task_videochat_tpo}     & 42.6 & 26.3 & 13.0 & 27.6 \\
VideoChatR1.5-7B~\citep{yan2026videochat}  & 52.4 & 32.3 & 16.8 & 35.5 \\

\midrule
\textbf{Ours}
& \textbf{63.7}
& \textbf{41.4}
& \textbf{23.9}
& \textbf{44.6} \\

\bottomrule
\end{tabular}%
}

\captionof{table}{
\textbf{Temporal grounding on ActivityNet.}
We compare all models in the zero-shot setting, evaluating temporal localization accuracy across multiple IoU thresholds and overall mean performance.}
\label{tab:activitynet}

\end{minipage}

\end{table*}
\textbf{i) Zero-shot Temporal Grounding.}\;
We first examine whether the temporal localization capability learned from STVG generalizes to two unseen temporal grounding benchmarks with distinct video distributions. Charades-STA~\citep{gao2017tall} primarily contains short indoor videos with densely occurring human activities, whereas ActivityNet~\citep{krishna2017dense_anet} contains longer videos covering a broader range of activities. We compare our model with prior methods evaluated under the same zero-shot setting. As shown in \Cref{tab:charades_sta} and \Cref{tab:activitynet}, our model demonstrates strong performance on both datasets. On Charades-STA, it improves the mean performance over the strongest prior zero-shot method by $+4.7$ points. The advantage is more pronounced on ActivityNet, where our model surpasses the strongest zero-shot baseline by $+9.1$ points in $m_{\mathrm{IoU}}$. Notably, the gains remain consistent under stricter overlap criteria, indicating that the model not only retrieves the relevant temporal region but also predicts more precise event boundaries. 

\begin{table*}[t]
\centering
\begin{minipage}[t]{0.43\textwidth}
\vspace{0pt}
\centering
\small
\setlength{\tabcolsep}{2.5pt}

\resizebox{\linewidth}{!}{%
\begin{tabular}{@{}lccccc@{}}
\toprule
& \multicolumn{5}{c}{\textbf{ReXTime}} \\
\cmidrule(lr){2-6}

\textbf{Model}
& \multicolumn{2}{c}{\textbf{Acc}}
& \multicolumn{3}{c}{\textbf{IoU}} \\

\cmidrule(lr){2-3}
\cmidrule(lr){4-6}

&
\textbf{Overall}
& $\mathbf{@0.5}$
& \textbf{Mean}
& $\mathbf{@0.3}$
& $\mathbf{@0.5}$ \\

\midrule
\multicolumn{6}{l}{{\scriptsize\linkblue{\textit{Finetuned}}}} \\

{TimeChat}~\citep{ren2024timechat}     
& \textcolor{darkgray}{49.4}
& \textcolor{darkgray}{11.1}
& \textcolor{darkgray}{26.5}
& \textcolor{darkgray}{40.5}
& \textcolor{darkgray}{21.9} \\

{VTimeLLM}~\citep{huang2024vtimellm}
& \textcolor{darkgray}{58.2}
& \textcolor{darkgray}{18.3}
& \textcolor{darkgray}{30.0}
& \textcolor{darkgray}{44.1}
& \textcolor{darkgray}{26.6} \\

{GraphThinker}~\citep{cheng2026graphthinker}
& \textcolor{darkgray}{71.3}
& \textcolor{darkgray}{30.8}
& \textcolor{darkgray}{41.5}
& \textcolor{darkgray}{57.5}
& \textcolor{darkgray}{40.4} \\

{VideoChat-R1.5}~\citep{yan2026videochat}
& \textcolor{darkgray}{74.8}
& \textcolor{darkgray}{38.1}
& \textcolor{darkgray}{45.8}
& \textcolor{darkgray}{61.8}
& \textcolor{darkgray}{46.4} \\

\midrule
\multicolumn{6}{l}{{\scriptsize\linkblue{\textit{Zero-Shot}}}} \\

VTimeLLM~\citep{huang2024vtimellm}
& 36.2 & -- & 20.1 & 28.8 & 17.4 \\

GraphThinker~\citep{cheng2026graphthinker}
& 66.8 & 15.2 & 25.3 & 33.9 & 20.3 \\

VideoTG-R1~\citep{dong2026videotg}
& 75.7 & 25.9 & 32.2 & 41.2 & 31.7 \\

\midrule
\textbf{Ours (ZS)}
& \underline{73.1}
& \textbf{43.0}
& \textbf{47.8}
& \textbf{62.1}
& \textbf{49.2} \\

\bottomrule
\end{tabular}%
}

\captionsetup{
    width=\linewidth,
    justification=justified,
    singlelinecheck=false
}
\captionof{table}{
\textbf{Grounded VideoQA on ReXTime.}
We evaluate on ReXTime~\citep{chen2024rextime}, which requires models to answer questions while localizing the temporal evidence supporting each prediction to measure joint reasoning and temporal localization.
}
\label{tab:rextime_results}

\end{minipage}
\hspace{0.02\textwidth}
\begin{minipage}[t]{0.53\textwidth}
\vspace{0pt}
\centering
\small
\setlength{\tabcolsep}{2.5pt}

\resizebox{\linewidth}{!}{%
\begin{tabular}{@{}lccc@{}}
\toprule
& \multicolumn{3}{c}{\textbf{Video Object Tracking}} \\
\cmidrule(lr){2-4}

\textbf{Model}
& \textbf{Ref-DAVIS}
& \textbf{Ref-YT-VOS}
& \textbf{ReasonVOS} \\

\midrule
\multicolumn{4}{l}{{\scriptsize\linkblue{\textit{Finetuned}}}} \\

{Molmo~\citep{deitke2025molmo} + SAM2~\citep{ravi2025sam}}
& \textcolor{darkgray}{65.2}
& \textcolor{darkgray}{--}
& 45.7 \\

{Sa2VA-Qwen3VL-4B}~\citep{yuan2025sa2va}
& \textcolor{darkgray}{76.0}
& \textcolor{darkgray}{68.1}
& 50.0 \\

{VideoGLaMM-7B}~\citep{munasinghe2025videoglamm}
& \textcolor{darkgray}{69.5}
& \textcolor{darkgray}{66.8}
& 33.9 \\

{VideoMolmo-7B}~\citep{ahmad2025videomolmo}
& \textcolor{darkgray}{72.5}
& \textcolor{darkgray}{67.3}
& 51.1 \\

{Molmo2-4B}~\citep{clark2026molmo2}
& \textcolor{darkgray}{73.5}
& \textcolor{darkgray}{70.2}
& 61.9 \\

\midrule
\multicolumn{4}{l}{{\scriptsize\linkblue{\textit{Zero-Shot}}}} \\

Qwen3-VL-4B~\citep{bai2025qwen3vl}
& 44.4 & 32.1 & 26.5 \\

Qwen3-VL-8B~\citep{bai2025qwen3vl}
& 41.0 & 48.3 & 24.9 \\

\midrule
\textbf{Ours (ZS) + SAM2}
& 81.9
& 72.8
& 62.1 \\

\textbf{Ours (ZS) + SAM3}
& \textbf{82.9}
& \textbf{74.9}
& \textbf{64.7} \\
\bottomrule
\end{tabular}%
}

\captionsetup{
    width=\linewidth,
    justification=justified,
    singlelinecheck=false
}
\captionof{table}{
\textbf{Video object tracking.}
We evaluate on referring video object segmentation benchmarks~\citep{khoreva2018video_refdavis, seo2020urvos_ytvos,bai2024one_reasonvos} using the $\mathcal{J}\&\mathcal{F}$ metric. Gray entries denote finetuned results, while ReasonVOS is evaluated zero-shot.
}
\label{tab:video_object_tracking}

\end{minipage}
\end{table*}
\textbf{ii) Zero-shot Grounded VideoQA.}\;
We further evaluate zero-shot generalization on ReXTime~\citep{chen2024rextime}, where the model is required to answer each question and provide supporting evidence through temporal localization. As shown in \Cref{tab:rextime_results}, our model demonstrates strong performance despite receiving no task-specific supervision. A key advantage is its temporal grounding accuracy: compared with the strongest zero-shot baseline, our model improves $m_{\mathrm{IoU}}$ by $+15.6$ points, while maintaining competitive answer accuracy. Notably, it also surpasses the finetuned VideoChat-R1.5~\citep{yan2026videochat} in $m_{\mathrm{IoU}}$ by $+2.0$ points. These results show that our model seamlessly supports evidence-aware video question answering without additional finetuning.

\textbf{iii) Zero-shot Video Object Tracking.}\;
We further examine whether the spatial grounding capability learned from STVG generalizes to referring video object tracking, where the model must maintain consistent localization of the queried object throughout the video. As shown in \Cref{tab:video_object_tracking}, our Qwen3-VL-4B-based model demonstrates strong zero-shot performance on Ref-DAVIS~\citep{khoreva2018video_refdavis}, Ref-YT-VOS~\citep{seo2020urvos_ytvos}, and ReasonVOS~\citep{bai2024one_reasonvos} when paired with off-the-shelf segmentation models. For a controlled comparison, Molmo2-4B~\citep{clark2026molmo2} similarly relies on SAM2~\citep{ravi2025sam} to propagate object masks, but provides point prompts, whereas our model generates bounding boxes. Using the same SAM2 backend, our model surpasses Molmo2-4B by $+8.4$, $+2.6$, and $+0.2$ points on the three benchmarks, respectively, despite the latter being finetuned on Ref-DAVIS and Ref-YT-VOS. Further, using SAM3~\citep{carion2026sam} consistently improves performance across all three datasets.

\subsection{Ablation Studies}
\label{sec:exp_ablation}
\begin{table*}[t]
\centering
\small
\setlength{\tabcolsep}{6pt}

\resizebox{\textwidth}{!}{%
\begin{tabular}{lcccccccc}
\toprule
&
\multicolumn{4}{c}{\textbf{Declarative Sentences}}
& \multicolumn{4}{c}{\textbf{Interrogative Sentences}} \\
\cmidrule(lr){2-5}
\cmidrule(lr){6-9}

\textbf{Reward}
& $m_{\mathrm{tIoU}}$
& $m_{\mathrm{vIoU}}$
& $\mathrm{vIoU}@0.3$
& $\mathrm{vIoU}@0.5$
& $m_{\mathrm{tIoU}}$
& $m_{\mathrm{vIoU}}$
& $\mathrm{vIoU}@0.3$
& $\mathrm{vIoU}@0.5$ \\

\midrule
SFT
& 50.0 & 34.9 & 48.5 & 33.3 & 
48.2 & 28.7 & 40.2 & 25.6 \\
Temporal
& 53.8 & 37.4 & 51.9 & 36.4
& 52.1 & 31.1 & 43.3 & 28.0 \\

Spatial 
& 51.8 & 37.2 & 51.6 & 35.9
& 49.9 & 31.3 & 43.2 & 29.2 \\

\textbf{Temporal + Spatial}
& \textbf{53.7}
& \textbf{38.3}
& \textbf{53.0}
& \textbf{37.5}
& \textbf{52.2}
& \textbf{32.4}
& \textbf{44.9}
& \textbf{30.3} \\

\bottomrule
\end{tabular}%
}

\caption{
Ablation of the reward design for GRPO on \textbf{VidSTG}.
We compare temporal-only, spatial-only, and the full temporal-spatial reward under declarative and interrogative settings.
All variants use Qwen3-VL-4B with Parallel Tube Decoding.
}
\label{tab:grpo_reward_ablation}
\end{table*}
\textbf{Reward design.}\; In \Cref{tab:grpo_reward_ablation}, we examine how the temporal and spatial rewards contribute to localization-aware policy optimization. With the temporal reward, we observe that the model corrects two common failure patterns: temporally displaced predictions and incomplete event coverage, where only the most salient part of an interaction is localized. Recovering a better-aligned event interval also improves $m_{\mathrm{vIoU}}$, since accurate spatial predictions can only contribute to the tube when they are generated over the correct temporal interval. Spatial reward addresses complementary localization errors in which the predicted boxes do not tightly align with the referred entity, include excessive surrounding context, or lose target consistency across frames. Combining both rewards achieves the strongest overall accuracy. We perform GRPO using outcome-level temporal and spatial rewards without introducing thinking tokens to avoid the additional decoding overhead. 

\section{Limitations}
\begin{figure}[t]
    \centering
    \includegraphics[width=0.99\columnwidth]{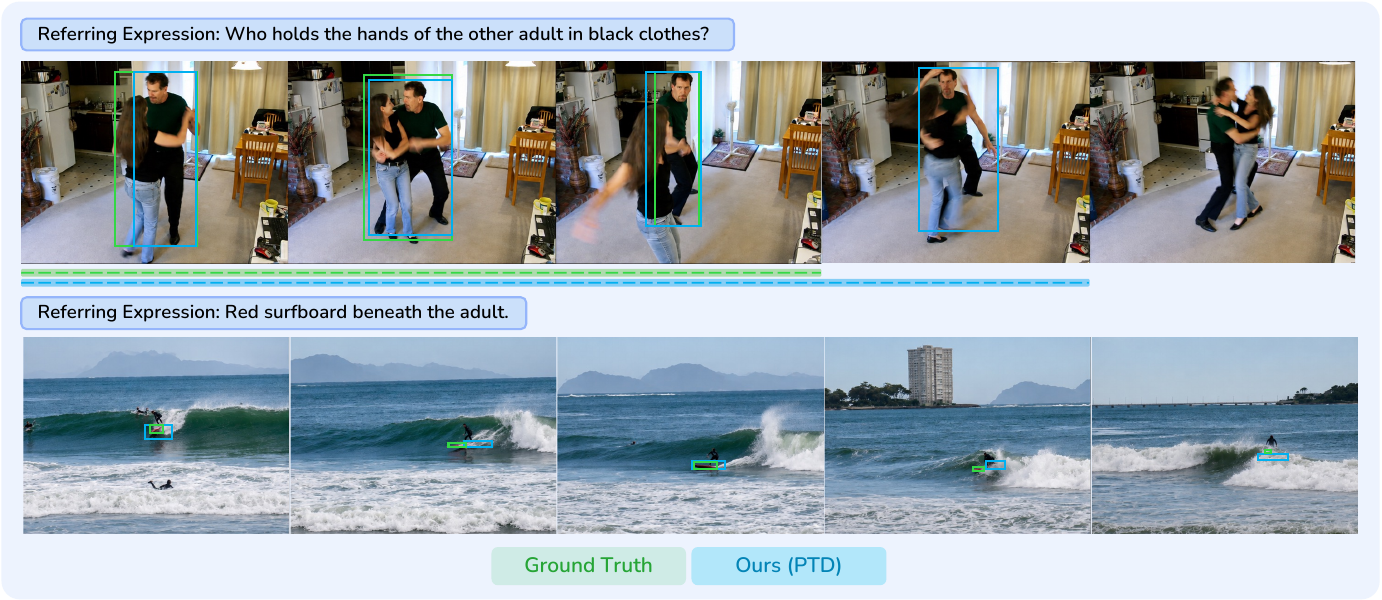}
    \caption{\textbf{Representative failure cases.} Temporally subtle state changes can produce ambiguous event boundaries (top), while spatial localization becomes difficult for small, rapidly moving, or occluded targets (bottom). Green and blue denote the ground-truth and PTD predictions, respectively; the horizontal lines indicate their temporal intervals.}
    \label{fig:failure_cases}
\end{figure} 
Our qualitative analysis in \Cref{fig:failure_cases} reveals two representative sources of error. We observe that spatial localization becomes more difficult when target identity must be recovered after occlusion or when the target is small and rapidly moving, which can lead to imprecise boxes (bottom). For temporal localization, failures are concentrated around short state transitions with visually ambiguous boundaries, causing the predicted interval to extend beyond or end before the complete event (top).

Beyond these failure cases, the current study remains bounded by the scope of existing STVG formulations and training resources. Standard STVG benchmarks represent each query using a single continuous temporal interval and one spatial tube, leaving multiple disjoint event occurrences, temporary out-of-view periods caused by camera motion, and multiple instances satisfying the same referring expression largely unexplored. Supporting these scenarios is important for real-world applications and motivates extending PTD toward multi-segment and multi-instance grounding. Moreover, large-scale STVG training data with dense tube annotations remains limited, with VidSTG~\citep{zhang2020does_vidstg_data} and HC-STVG-v1/v2~\citep{tang2021human_hcstvg_data} serving as the primary datasets available for this setting. In future work, we plan to expand the scope toward more generalized grounding settings by supporting richer multi-segment and multi-instance outputs while diversifying the training data across broader video domains, entities, events, and interaction types.
\section{Conclusion}
We introduced Parallel Tube Decoding (PTD), a generative formulation for spatio-temporal video grounding that removes the need to decode dense spatial trajectories autoregressively. By first predicting the relevant temporal interval and then generating all time-conditioned spatial blocks in parallel, PTD reduces tube generation to just two sequential decoding rounds, independent of tube length. Decoupled Block Attention further removes dependencies between spatial predictions while preserving access to the shared multimodal evidence, and localization-aware policy optimization improves temporal boundaries and bounding-box geometry.

Our experiments show that removing these sequential dependencies improves both efficiency and localization quality. On VidSTG, PTD reduces Tube Completion Latency by 79× and increases spatial decoding throughput by 92× relative to standard autoregressive decoding, while achieving stronger grounding accuracy. Using a 4B backbone, the resulting model performs favorably well on VidSTG and HC-STVG, despite relying on no dedicated spatial localization module. The learned grounding capability also transfers zero-shot to temporal grounding, grounded VideoQA, and referring video object tracking. Overall, our results show that dense spatio-temporal localization can be generated efficiently without sequential trajectory decoding. 
\clearpage
\bibliographystyle{assets/plainnat}
\bibliography{resources/main}

\end{document}